\documentclass[runningheads]{llncs}
\usepackage[T1]{fontenc}
\usepackage{graphicx}

\usepackage{amsmath,amsfonts}
\usepackage{algorithmicx}
\usepackage{algpseudocode}
\usepackage{algorithm}
\usepackage{graphicx}
\usepackage{textcomp}
\usepackage{xcolor}
\usepackage{array}
\usepackage{xspace}
\usepackage{enumitem}
\usepackage{pifont}
\usepackage{multirow}
\usepackage{tabularx}
\usepackage[normalem]{ulem}

\newcommand{\code}[2]{\textcolor{#1}{\texttt{#2}}}

\newcommand{\vs}{\textit{v}.\textit{s}.\xspace}

\usepackage{xspace}
\newcommand{\xname}{AutoPPA\xspace}

\newcommand{\EEI}{$E^2 I$\xspace}

\newcommand{\rtlrewriter}{RTLRewriter\xspace}
\newcommand{\symrtlo}{SymRTLO\xspace}

\newcommand{\adp}[3]{%
  {#3\% }{\scriptsize/\, #2\% /\, #1\%}%
}

\newcommand{\dap}[3]{%
  {#3\% }{\scriptsize/\, #1\% /\, #2\%}%
}

\newcommand{\dcadp}[3]{%
  {#1 }{\scriptsize/\, #2 /\, #3}%
}

\begin{document}

\title{\xname: Automated Circuit PPA Optimization via
Contrastive Code-based Rule Library Learning}

\titlerunning{\xname}

\author{Chongxiao Li\inst{1,2} \and
Pengwei Jin\inst{1} \and
Di Huang\inst{1} \and
Guangrun Sun\inst{1,2} \and
Husheng Han\inst{1} \and
Jianan Mu\inst{1} \and
Xinyao Zheng\inst{1,2} \and
Jiaguo Zhu\inst{1,3} \and
Shuyi Xing\inst{1,3} \and
Hanjun Wei\inst{1,2} \and
Tianyun Ma\inst{4} \and
Shuyao Cheng\inst{1} \and
Rui Zhang\inst{1} \and
Ying Wang\inst{1} \and
Zidong Du\inst{1} \and
Qi Guo\inst{1} \and
Xing Hu\inst{1}
}

\authorrunning{C. Li et al.}

\institute{State Key Lab of Processors, Institute of Computing Technology, Chinese Academy of Sciences, Beijing, China \and University of Chinese Academy of Sciences, Beijing, China \and University of Science and Technology of China, Hefei, China \and Institute of AI for Industries, Chinese Academy of Sciences, Nanjing, China}

\maketitle              

\begin{abstract}

Performance, power, and area (PPA) optimization is a fundamental task in RTL design, requiring a precise understanding of circuit functionality and the relationship between circuit structures and PPA metrics. Recent studies attempt to automate this process using LLMs, but neither feedback-based nor knowledge-based methods are efficient enough, as they either design without any prior knowledge or rely heavily on human-summarized optimization rules.

In this paper, we propose \xname, a fully automated PPA optimization framework. The key idea is to automatically generate optimization rules that enhance the search for optimal solutions. 
To do this, \xname employs an \textit{Explore-Evaluate-Induce} (\EEI) workflow that contrasts and abstracts rules from diverse generated code pairs rather than manually defined prior knowledge, yielding better optimization patterns.
To make the abstracted rules more generalizable, \xname employs an adaptive multi-step search framework that adopts the most effective rules for a given circuit.
Experiments show that \xname outperforms both the manual optimization and the state-of-the-art methods \symrtlo and \rtlrewriter.

\end{abstract}

\section{Introduction}

Performance, power, and area (PPA) constitute the fundamental metrics for evaluating integrated circuit design quality. As a critical design challenge, PPA optimization demands substantial expertise in hardware implementation, especially in RTL design, because it requires a precise understanding of circuit functionality and how circuit structures affect post-synthesis PPA results. 

Recently, some large-language-model(LLM)-based efforts have attempted to automate this process. These studies fall into two categories: 1) Direct feedback methods that utilize post-synthesis PPA metrics as LLM inputs~\cite{chipgpt,thorat2024advancedlargelanguagemodel}. While straightforward, these approaches demonstrate limited efficacy because LLMs lack an understanding of the correlations between circuit structure and PPA metrics. 
2) Knowledge-based methods that employ manually curated or human-summarized PPA optimization rules~\cite{yao2024rtlrewriter,wang2025symrtlo}. Although potentially more targeted, these solutions face scalability challenges, for the labor-intensive knowledge base construction inherently limits both the coverage of circuit patterns and the diversity of optimization techniques, consequently limiting their practical utility. Even in some commercial tool guidebooks, only a few dozen sample entries are provided~\cite{synopsys2012datapath}.

\begin{figure}[t]
  \centering
\includegraphics[width=\linewidth]{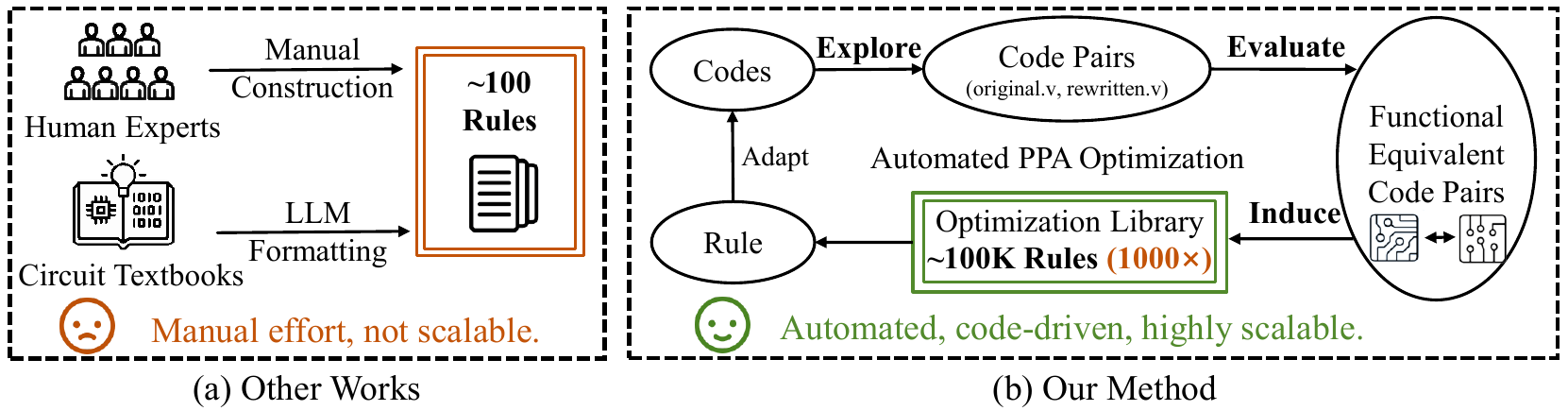}
\caption{Constructing the Rule Library for PPA optimization.}
\label{fig:insight}
\end{figure}

In this paper, to address the scalability problem in knowledge-based optimization, we investigate the fundamental problem: \textit{Can we synthesize reusable PPA optimization knowledge automatically, without any human intervention, from just raw RTL code?} We observe that, leveraging the strong code-generation capabilities of LLMs, it is possible to rewrite functionally equivalent RTL codes with diverse structures. As shown in Figure~\ref{fig:insight}, these contrastive code pairs can be used to induce the optimization rule library, without relying on manual rules.

However, synthesizing PPA optimization knowledge automatically remains critically challenging due to the following reasons: 
\textbf{Challenge 1: the evaluation of optimization rules.}  Assessing rewritten RTL is inherently multifaceted. First, any candidate that is not functionally equivalent is unusable, so robust equivalence checking between any original RTL and its rewritten ones is essential. Second, beyond functional correctness, evaluation must balance PPA with the diversity of RTL samples.
\textbf{Challenge 2: the high-quality rule induction.}  Even when optimized code pairs are available, directly extracting high-quality rules is difficult. Naive summarization produces rules cluttered with low-level, implementation-specific idiosyncrasies, and different code pairs can produce redundant or contradictory rules. Such noisy rule sets are hard to apply and lead to suboptimal optimization in practice. 
\textbf{Challenge 3: the abstraction gap between rules and designs.} Overly specific rules introduce retrieval noise, while excessively abstract rules lack optimization value and can hardly benefit PPA results in practical usage. With a large-scale, automatically generated rule library, efficiently incorporating rules for PPA optimization is challenging because of the gap between abstract rules and specific designs. The rules provide general circuit descriptions, making it difficult for LLMs to locate relevant RTL code snippets and implement compliant modifications, resulting in functionally inequivalent or suboptimal results.

\begin{figure*}[t]
  \centering
\includegraphics[width=\linewidth]{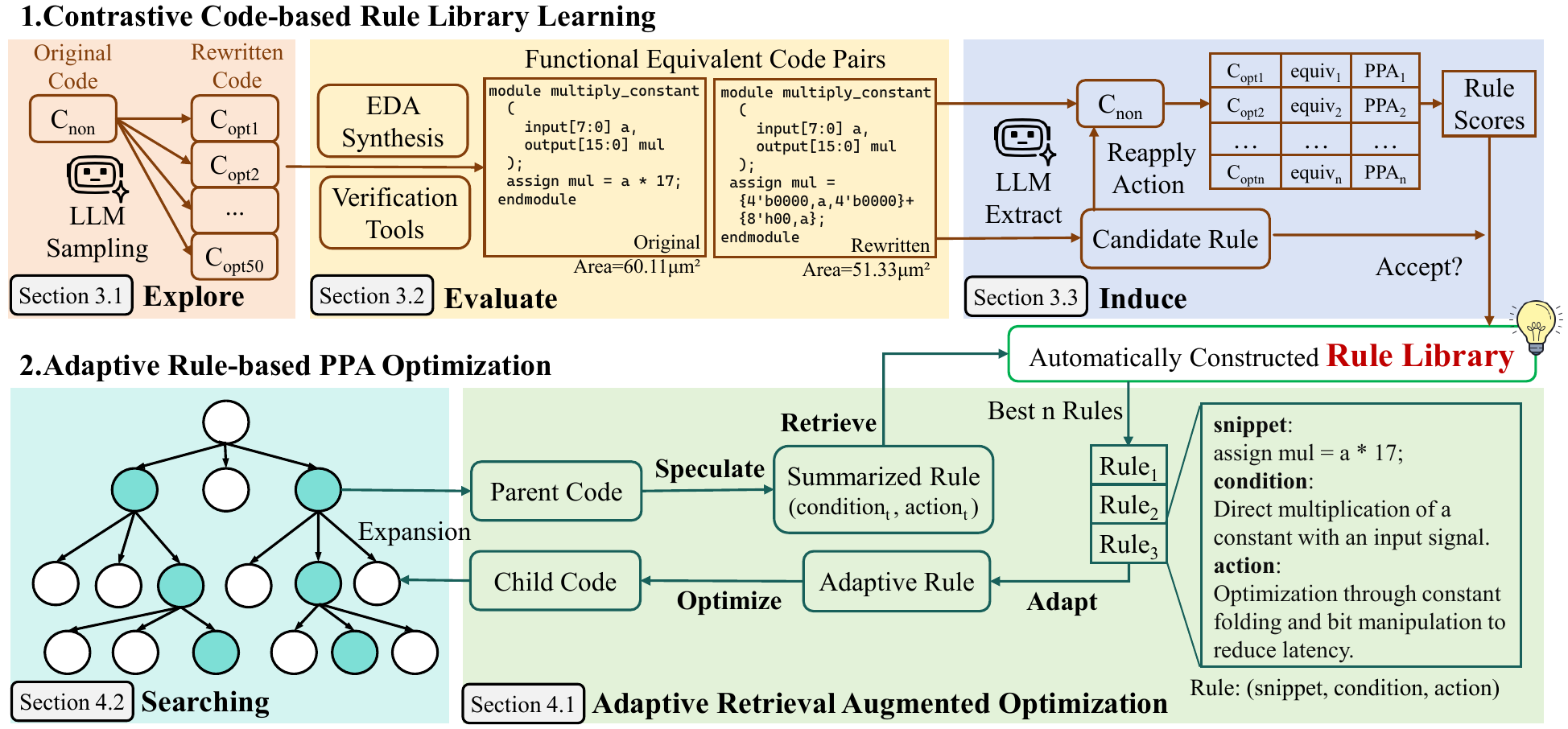}
\caption{Overview of \xname. \xname includes the pipeline of the rule library learning workflow and the adaptive rule-based PPA optimization.}
\label{fig:method_overview}
\end{figure*}

To this end, we propose a fully automated PPA optimization framework, \xname, which includes an \textit{Explore-Evaluate-Induce} (\EEI) workflow and an adaptive multi-step search method.
As shown in Figure ~\ref{fig:insight}(b), the \EEI workflow (1) \textbf{explores} the Verilog code pairs via multiple rounds of random LLM rewriting sampling; (2) \textbf{evaluates} and verifies the functional equivalence between the original code and the rewritten code, and builds functional-equivalent Verilog code pairs via verification tools against \textbf{Challenge 1}; and (3) \textbf{induces}   optimization rules with the form of (\textit{snippet}, \textit{condition}, \textit{action}) triples by analysing paired code snippets and their PPA labels against \textbf{Challenge 2}.
The adaptive multi-step search framework is a rule-enhanced beam search method that leverages the rule to better guide the LLM's exploration of higher-quality Verilog code samples, increasing the probability of PPA-optimized implementations
against \textbf{Challenge 3}.

Experiments show that \xname achieves a maximum of 15.31\% area improvements and 11.28\% delay improvements on 60 comprehensive benchmark circuits, and surpasses the area optimization results of manual optimization by 19.25\% and state-of-the-art by 7.56\% on 11 representative circuits, respectively.

\section{Overview}

\subsection{Background: RTL PPA Optimization}

From a circuit-quality perspective, integrated-circuit (IC) design proceeds through three main stages~\cite{huang2021ml4edasurvey,pan2025surveyllm4eda}: At the Register-Transfer Level (RTL), engineers specify functionality in Verilog or VHDL; Automated logic synthesis then converts the RTL description into a gate-level netlist; Finally, place-and-route tools in the physical-design stage produce the final layout while optimizing timing, power, and area. Although each stage affects PPA (performance, power, and area), RTL, sitting at the top of the design hierarchy, has the greatest influence: redundant or suboptimal RTL limits what downstream synthesis and physical tools can recover. Consequently, producing high-quality RTL is essential to meet stringent PPA targets. Despite advances in automation, RTL PPA optimization still relies heavily on expertise and manual tuning, as it is time-consuming and its outcomes can vary substantially with the designer’s experience.

\subsection{\xname}
As shown in Figure~\ref{fig:method_overview}, \xname consists of two phases: 1. Contrastive Code-based Rule Library Learning (Section~\ref{sec:Rule Library Learning}), and 2. Adaptive Rule-based PPA Optimization (Section~\ref{sec:rule_based_opt}).

Contrastive Code-based Rule Library Learning is proposed to automatically construct a comprehensive rule library without any human intervention. 
It is composed of a \textit{Explore-Evaluate-Induce} workflow, where we first explore optimization opportunities through code pair sampling, then evaluate the code pairs based on their equivalence and PPA, and finally induce natural language optimization rules from the code pairs.

Adaptive Rule-based PPA Optimization is a novel retrieval-augmented optimization approach proposed to exploit the learned rule library. It integrates single-step adaptive retrieval with multi-step rule-based enhanced search, enabling automatic selection and combination of the most effective rules for circuit PPA optimization.

\section{Contrastive Code-based Rule Library Learning}\label{sec:Rule Library Learning}
Contrastive Code-based Rule Library Learning is proposed to construct the rule library from raw RTL code without any human intervention. It consists of three stages, \textit{Explore-Evaluate-Induce}, whose details are shown as follows.

\subsection{Explore: Code Sampling} 
\label{subsec:explore}
In this stage, we aim to create a sufficiently large and diverse corpus of functionally valid RTL code base from which meaningful optimization rules can emerge.
Approximately 130K Verilog RTL codes are crawled from public GitHub repositories. These codes undergo a two-phase synthesis process:
\begin{itemize}[label={}, leftmargin=0pt]
    \item \textbf{Lightweight Filtering:} We initially synthesize all collected designs using the lightweight \texttt{Yosys} tool~\cite{yosys} to discard any non-self-contained or non-synthesizable designs.
    \item \textbf{Accurate Synthesis and PPA Metrics:} The remaining filtered codes are then synthesized using \code{black}{SiliconCompiler}~\cite{siliconcompiler}, which combines \texttt{Yosys} and \texttt{OpenSTA}~\cite{ajayi2019openroad} to provide more accurate PPA measurements at the cost of increased synthesis time. Both the RTL code and the associated PPA performance metrics are stored for subsequent rule library learning.
\end{itemize}

With a large corpus of original circuit code, we explore diverse rewrite forms of the initial designs. General-purpose language models lack accurate and specialized knowledge for circuit optimization but show strong ability to sample massive and varied code. We leverage this ability to generate multiple rewrites of the initial circuits for potential optimization insight discovery. We first select 100K RTL codes with mid-range area values from the filtered corpus. Each code is rewritten N (default N = 50) times by the LLM Qwen2.5-Coder-7B-Instruct~\cite{hui2024qwen25codertechnicalreport}, producing over 5M pairs of circuit codes. These extensively explored pairs are subsequently passed to the evaluation stage to identify and retain those that may contain valuable optimization insights.

\subsection{Evaluate: Contrastive Code Pairs Evaluation Using EDA Tools}
\label{subsec:evaluate_contrastive_pairs}

In this stage, we aim to evaluate the massive code pairs, identifying functionally equivalent pairs that exhibit significant differences in PPA through evaluation, laying the groundwork for extracting robust optimization rules. After obtaining a large number of code pairs with diverse rewrite patterns, we first evaluate their PPA metrics and functional equivalence.

For PPA evaluation, we adopt the same settings as in Section~\ref{subsec:explore}. Functional equivalence evaluation is more challenging because it is difficult to construct testbenches for circuits with diverse trigger patterns, timing behaviors, and functionalities. To address this, we develop an automatic testbench generator. Using \code{black}{Yosys}, we automatically extract the top module, port, clock and reset information of each circuit. We then construct a testbench for all code pairs. The testbench fully verifies clock and reset behaviors, applies multiple long random stimulus sequences, and compares the outputs of the original and rewritten circuits to verify equivalence. The generated testbenches achieve 100\% line coverage and high reliability on our test circuits.

After verifying and synthesizing all rewrites of each RTL design, we compute the Shannon entropy of the PPA distribution among functionally equivalent versions to measure diversity. A higher entropy value indicates greater diversity in possible optimization patterns that can be extracted. Non-equivalent codes are given zero PPA improvement and excluded from the calculation. We rank designs by entropy and select the top \(K\%\) (\(K=25\)) as candidates. For each selected design, we keep only equivalent code pairs with relative PPA differences above 5\%, denoted as \((C_{non}, C_{opt})\) for non-optimized and optimized versions.

\subsection{Induce: Contrastive-Code Based Rule Induction}
\label{subsec:Induce}

In this stage, we summarize key optimization insights into higher-level rules.
To extract actionable circuit optimization strategies, we define each rule as a triplet (\textit{snippet}, \textit{condition}, \textit{action}):
\begin{itemize}[label={}, leftmargin=0pt]
    \item \textbf{\textit{snippet}}: Representative low-efficiency RTL code snippets that serve as optimization targets.
    \item \textbf{\textit{condition}}: Generalized contextual conditions describing when the optimization rule becomes applicable.
    \item \textbf{\textit{action}}: Transformation actions specifying how to modify original RTL codes to achieve optimized PPA.
\end{itemize}
For each pair $(C_{non}, C_{opt})$, we first generate a set of candidate rules via the LLM, leading to $n$ (default $n$ = 2) distinct rules that might explain the improvement observed in $C_{opt}$. 

We then identify high-quality rules through systematic evaluation. Each rule is reapplied to the original circuit with multiple LLM optimization attempts, and compared with the original non-optimized and optimized circuit PPA in the code pair. We normalize outcomes using Eq~\ref{eq:rewr_score}:
\begin{equation}\label{eq:rewr_score}
s_i=\mathbf{1}_{\text{eq}}(i)\;\cdot\;\mathrm{clip}_{[0,1]}\!\left(\alpha+\beta\,\frac{\mathrm{PPA}_n - \mathrm{PPA}_i}{\mathrm{PPA}_n - \mathrm{PPA}_o}\right)
\end{equation}
where $s_i$ denotes the score of the $i$‑th rewritten circuit. $\mathrm{PPA}_n$, $\mathrm{PPA}_o$, and $\mathrm{PPA}_i$ are the PPA values of the non‑optimized circuit $C_{non}$, the optimized circuit $C_{opt}$, and the $i$‑th rewrite, respectively. $\mathbf{1}_{\text{eq}}(i)$ ensures that only functionally equivalent rewrites receive nonzero scores. The parameters $\alpha$ (default 0.25) and $\beta$ (default 0.5) normalize the improvement: an equivalent rewrite with no PPA gain receives a score of 0.25, while one matching the optimized circuit’s improvement receives 0.75.

Scores are then averaged across all attempts per rule. Rules with average scores above 0.7 are classified as high-quality and added into the rule library, indicating strong circuit improvement potential.

\section{Adaptive Rule-based PPA Optimization}\label{sec:rule_based_opt}

To fully leverage the optimization rule library in practical scenarios, we propose an integrated framework called \textit{Adaptive Rule-based PPA Optimization}. 
This framework integrates single-step Adaptive Retrieval Augmented Optimization (ARAO) with multi-step Rule-based Enhanced Searching. The single-step optimization extracts structural conditions from the input RTL code and applies the most adaptive actions to the circuits based on rules,  while the multi-step searching iteratively refines candidates via beam search, balancing both PPA gains and circuit diversity.

\subsection{Adaptive Retrieval Augmented Optimization}\label{subsec:mthd_ARAO}
To harness the optimization rule library for RTL code PPA optimization, we retrieve rules by matching structural conditions and potential actions in the RTL code, and use the retrieved rules to guide LLMs to optimize the RTL code. We observe that some rules contain descriptions targeting their source code-pairs, which may contaminate the optimization context of the current circuit. We also observe that direct retrieval by semantic embedding similarity may introduce noise. Therefore, we design an Adaptive Retrieval Augmented Optimization (ARAO) process that reduces the impact of these issues. This process contains four steps:
\begin{itemize}[label={}, leftmargin=0pt]
    \item \textbf{Speculate:} The LLM is prompted to summarize an optimization rule ($\textit{snippet}_t$, $\textit{condition}_t$, $\textit{action}_t$) for the target code.
    \item \textbf{Retrieve:} Using the $\textit{condition}_t$ and $\textit{action}_t$, we retrieve the three most similar rules from the rule library based on cosine similarity between rule embeddings.
    \item \textbf{Adapt:} To mitigate interference from source-specific information in retrieved rules, we utilize the LLM to adapt these rules with the target code, ensuring their applicability and relevance.
    \item \textbf{Optimize:} The target code and adaptive rules are input to the LLM, which generates optimized code variants. 
    These variants are then verified for functional equivalence and synthesized to assess PPA improvements.
\end{itemize}

\subsection{Rule-based Enhanced Searching}\label{subsec:mthd_kes}
Although ARAO is effective for single-step optimization, comprehensive PPA improvement often requires efficient search in a large search space of complex code optimization. 
We thus propose a multi-step search framework based on beam search.
At each iteration, the top-$k$ scored candidate codes (beam width $k$) are selected.
For each, the ARAO process extends $m$ optimized variants. 
The next iteration's candidate pool comprises all new variants and the current top candidates, up to a maximum of $s$ iterations.

The candidate scoring function is defined as:
\begin{equation}\label{eq:search_score}
\small{
    \text{Score} = \omega \cdot \text{Score}_{diversity} + (1-\omega) \cdot \text{Score}_{ppa}}
\end{equation}
where $\text{Score}_{diversity}$ quantifies diversity using TF-IDF similarity~\cite{aizawa2003information} to parent codes, $\text{Score}_{ppa}$ incorporates functional equivalence and relative PPA performance, and $\omega=0.25$.
This encourages both PPA improvement and exploration of diverse optimization strategies.

\section{Experiments}

We first introduce our experimental setup in Section~\ref{subsec:exp_setting}. Then, we conduct comprehensive experiments and thorough ablation studies to validate the effectiveness of our approach in Section~\ref{subsec:exp_result}.

\subsection{Experimental Setup}
\label{subsec:exp_setting}

\textbf{Benchmarks.}
We use the RTLRewriter~\cite{yao2024rtlrewriter} benchmark, which is specifically designed for RTL code optimization. It contains 54 designs with comprehensive optimization patterns and 3 practical, large designs that are synthesizable. Following RTLRewriter, we partition these 3 practical designs into 6 distinct modules, resulting in a total of 60 designs.
For each design, we generate a rigorous testbench (100\% line and branch coverage on non-redundant code) to verify the equivalence between the optimized code and the original implementation.

\textbf{Baselines.}
\textbf{\textit{(1) SOTA RTL code optimization methods.}} To compare with \rtlrewriter, we synthesize all 11 optimized circuits from their public repository using our aligned synthesis flow.
For \symrtlo~\cite{wang2025symrtlo}, we compare our results with their self-reported results on 5 circuits that implement complex algorithms.
\textbf{\textit{(2) Proficient Verilog engineers}}. We compare our method with results manually optimized by a Verilog engineer with over two years of experience on the complete benchmark. The engineer is provided with standard verification and synthesis environments and allowed 16 working hours to perform thorough optimizations while summarizing optimization rules during this process.
\textbf{\textit{(3) Representative LLMs.}} We compare our results with representative LLMs, listed in Table~\ref{tab:baseline_llms}. For a fair comparison, we evaluate the most optimized code generated by each LLM under an identical amount of sampled RTL code. In all experiments, we use a unified generation temperature of 0.6 and provide prompts that are consistent with our approach.

\textbf{Metrics.}
We report the synthesized total cell area (in $\mu m^2$), cycle delay (in $ns$), and dynamic power (in $mW$) to evaluate circuit area, performance, and power. Unless otherwise specified, all syntheses use \code{black}{SiliconCompiler} with the FreePDK 45nm process~\cite{Stine2007freepdk}. For equivalent and optimized circuits, we report their PPA improvement according to $Impr=1- { PPA_{opt}}/{ PPA_{original}}$. If the generated circuits are inequivalent or with worse target PPA, we mark them as 0 improvement. For the vanilla LLM baseline, following the widely adopted $Pass@k$ metric~\cite{chen2021evaluatinglargelanguagemodels} in code generation tasks, we propose the \textbf{Impr@k} metric to estimate the expected optimal PPA optimization  from \(k\) samples, using the results from a total of \(n > k\) samples, as shown in Equation~\ref{eq:impr_at_k}:
\begin{equation}
\textstyle
\operatorname{Impr@k} := \mathbb{E}_{\text{circuits}}\left[
\sum_{j=1}^n \left( 
   Impr_j^{\downarrow} \cdot \frac{\binom{n-j}{k-1}}{\binom{n}{k}} 
\right)
\right]
\label{eq:impr_at_k}
\end{equation}
where the \({Impr}^{\downarrow}_j\) denotes the \(j\)-th largest PPA improvements in $n$ samples , i.e., \({Impr}^{\downarrow}_1 \ge {Impr}^{\downarrow}_2 \ge \cdots \ge {Impr}^{\downarrow}_n\).

\textbf{\xname settings.}
We configure different search scales according to Table~\ref{tab:search_setting}, generating 15, 50, 104, and 210 RTL code samples for each configuration. We utilize gte\_Qwen2-7B-instruct~\cite{li2023towards} as the embedding model to embed the rule library and retrieval queries. We configure \xname with both smaller Qwen2.5-7B-Instruct and Qwen2.5-Coder-7B-Instruct models and larger models DeepSeek-V3-0324 model during search to show the generality of our framework. We set the LLM generation temperature to 0.6 for all models.

\begin{table}[t]
\centering
\caption{Baseline language models in our experiments.}

\resizebox{0.7\linewidth}{!}{%
\begin{tabular}{|c|c|c|c|}
\hline
Type                          & Model                      & Abbr. & \# Params \\ \hline
\multirow{2}{*}{RTL Specific} & CodeV-All-QC              & CodeV        & 7.62B         \\
 & HaVen-DeepSeek  & HaVen & 6.74B \\ \hline
Coding Specific               & Qwen2.5-Coder-7B-Instruct  & Qwen Coder          & 7.62B         \\ \hline
Reasoning &
  DeepSeek-R1-Distill-Qwen-7B &
  DS-R1-Dist &
  7.62B \\ \hline
General Purpose                & DeepSeek-V3-0324         & DeepSeek-V3        & 685B          \\ \hline
\end{tabular}%
}
\label{tab:baseline_llms}

\end{table}

\begin{table}[htb]
\centering
\caption{Configurations for \xname's searching scales, where $n = (1 + k \cdot (s-1)) \cdot m$.}

\label{tab:search_setting}
\resizebox{0.7\columnwidth}{!}{%
\begin{tabular}{|c|ccc|c|}
\hline
\begin{tabular}[c]{@{}c@{}}Config\\ Abbr.\end{tabular} &
  \begin{tabular}[c]{@{}c@{}}Beam Width\\ ($k$)\end{tabular} &
  \begin{tabular}[c]{@{}c@{}}Num Expand \\ ($m$)\end{tabular} &
  \begin{tabular}[c]{@{}c@{}}Max Steps\\ ($s$)\end{tabular} &
  \begin{tabular}[c]{@{}c@{}}Total RTL Searched\\ ($n$)\end{tabular} \\ \hline
2-3-3  & 2 & 3  & 3 & 15  \\
3-5-4  & 3 & 5  & 4 & 50  \\
3-8-5  & 3 & 8  & 5 & 104 \\
5-10-5 & 5 & 10 & 5 & 210 \\ \hline
\end{tabular}%
}

\end{table}

\subsection{Experimental Results}
\label{subsec:exp_result}

\textbf{\xname surpasses manual effort and SOTA methods.} 
\textit{(1) \xname outperforms RTLRewriter and manual optimizations on area-oriented circuit optimizations.}
We conduct a circuit-by-circuit comparison against RTLRewriter's open-sourced optimized RTL code and manually optimized results.
We employ DeepSeek-V3 as \xname's backbone model with 5-10-5 search settings, and align the optimization target with RTLRewriter for area-oriented optimization. The results are shown in Table~\ref{tab:comp_with_rtlrewriter}, with the best results highlighted in bold and results that fail to achieve optimization compared to the original design shown in light color.
Our method achieves superior performance, obtaining the smallest area in 10 out of 11 circuits. On average, compared to RTLRewriter, our approach delivers significant improvements: 7.56\% in area and 9.00\% in power. The relatively poor performance of manual optimization demonstrates that optimizing these circuits presents considerable challenges even for human engineers with proficient experience.

\begin{table}[]
\caption{\uline{Area-oriented} optimization comparison of \xname with manual efforts and RTLRewriter. \xname outperforms manual efforts by 19.25\% and RTLRewriter by 7.56\%. }

\label{tab:comp_with_rtlrewriter}
\resizebox{\linewidth}{!}{%
\begin{tabular}{|c|c|c|c|c|}
\hline
\multirow{2}{*}{Design} &
  Original &
  \multicolumn{1}{c|}{Manual} &
  RTLRewriter &
  \xname-DS \\ \cline{2-5} 
 &
  \dcadp{\textbf{Area}}{Delay}{Power} &
  \multicolumn{1}{c|}{\dcadp{\textbf{Area}}{Delay}{Power}} &
  \dcadp{\textbf{Area}}{Delay}{Power} &
  \dcadp{\textbf{Area}}{Delay}{Power} \\ \hline
add3               & \dcadp{65.17}{0.26}{0.85} & \color[HTML]{9B9B9B}\dcadp{65.17}{0.26}{0.85} & \dcadp{\textbf{42.83}}{0.21}{0.62} & \color[HTML]{9B9B9B}\dcadp{65.17}{0.26}{0.85} \\
mux\_type1         & \dcadp{2.39}{0.04}{0.00010} & \color[HTML]{9B9B9B}\dcadp{2.39}{0.04}{0.00010} & \dcadp{\textbf{2.13}}{0.04}{0.00010} & \dcadp{\textbf{2.13}}{0.04}{0.00010} \\
mux\_type3         & \dcadp{2.39}{0.05}{0.00010} & \color[HTML]{9B9B9B}\dcadp{2.39}{0.05}{0.00010} & \color[HTML]{9B9B9B}\dcadp{2.39}{0.05}{0.00010} & \color[HTML]{9B9B9B}\dcadp{2.39}{0.05}{0.00010} \\
mux\_type5         & \dcadp{6.12}{0.13}{0.00010} & \color[HTML]{9B9B9B}\dcadp{6.12}{0.13}{0.00010} & \color[HTML]{9B9B9B}\dcadp{6.12}{0.13}{0.00010} & \color[HTML]{9B9B9B}\dcadp{6.12}{0.13}{0.00010} \\
example1           & \dcadp{61.71}{0.22}{0.78} & \dcadp{60.116}{0.21}{0.87} & \dcadp{48.15}{0.24}{0.84} & \dcadp{\textbf{29.26}}{0.23}{0.68} \\
example3           & \dcadp{52.40}{0.27}{2.33} & \dcadp{49.476}{0.31}{2.07} & \dcadp{37.51}{0.21}{1.54} & \dcadp{\textbf{36.44}}{0.21}{1.25} \\
com\_subexp       & \dcadp{11967.34}{2.98}{0.26} & \dcadp{11370.17}{3.00}{0.24} & \dcadp{11389.58}{3.12}{0.24} & \dcadp{\textbf{11310.85}}{3.06}{0.24} \\
add\_bit\_wid  & \dcadp{63.57}{0.32}{0.0016} & \color[HTML]{9B9B9B}\dcadp{63.57}{0.32}{0.0016} & \color[HTML]{9B9B9B}\dcadp{63.57}{0.32}{0.0016} & \dcadp{\textbf{36.44}}{0.59}{0.00070} \\
add\_subexp       & \dcadp{132.73}{0.57}{0.0029} & \color[HTML]{9B9B9B}\dcadp{132.73}{0.57}{0.0029} & \color[HTML]{9B9B9B}\dcadp{132.74}{0.57}{0.0029} & \color[HTML]{9B9B9B}\dcadp{132.73}{0.57}{0.0029} \\
m\_con\_mul  & \dcadp{1374.16}{1.17}{0.031} & \color[HTML]{9B9B9B}\dcadp{1374.16}{1.17}{0.031} & \dcadp{1026.76}{1.13}{0.023} & \dcadp{\textbf{876.47}}{1.04}{0.02} \\
m\_con\_mul2 & \dcadp{1628.72}{1.22}{0.038} & \color[HTML]{9B9B9B}\dcadp{1628.72}{1.22}{0.038} & \dcadp{1370.17}{1.22}{0.031} & \dcadp{\textbf{873.01}}{1.16}{0.021} \\ \hline
Avg. Impr.         & - & \dcadp{1.20\%}{-0.99\%}{0.60\%}  & \dcadp{12.89\%}{2.83\%}{9.35\%} & \dcadp{\textbf{20.45\%}}{-4.85\%}{18.35\%} \\ \hline
\end{tabular}%
}

\end{table}

\begin{table*}[t]
\centering
\caption{Comparison of the \uline{area-oriented} optimization results of \xname, manual optimization, vanilla DeepSeek-V3 sampling, and SymRTLO, synthesizing with commercial tools. Results for SymRTLO are copied from their original paper.}

\resizebox{0.85\linewidth}{!}{%
\begin{tabular}{|cc|ccc|}
\hline
\multicolumn{2}{|c|}{Design} & spmv & subexp\_elim & adder\_architecture \\ \hline
\multicolumn{2}{|c|}{Metric} & \dcadp{\textbf{Area}}{Delay}{Power} & \dcadp{\textbf{Area}}{Delay}{Power} & \dcadp{\textbf{Area}}{Delay}{Power} \\ \hline

\multicolumn{1}{|c|}{\multirow{2}{*}{SSC}} 
& Original    & \dcadp{22908.69}{7.95}{1.46} & \dcadp{9484.15}{11.78}{4.61} & \dcadp{541.92}{2.29}{0.17} \\
\multicolumn{1}{|c|}{} 
& SymRTLO     & {\color[HTML]{9B9B9B}\dcadp{22908.69}{7.95}{1.46}} & \dcadp{6791.88}{11.78}{3.53} & \dcadp{531.88}{2.48}{0.17} \\ \hline

\multicolumn{1}{|c|}{\multirow{4}{*}{12nm}} 
& Original    & \dcadp{423.94}{0.16}{6.71} & \dcadp{162.75}{1.27}{0.22} & \dcadp{8.48}{0.15}{0.01} \\
\multicolumn{1}{|c|}{} 
& Manual      & {\color[HTML]{9B9B9B}\dcadp{423.94}{0.16}{6.71}} & {\color[HTML]{9B9B9B}\dcadp{162.75}{1.27}{0.22}} & {\color[HTML]{9B9B9B}\dcadp{8.48}{0.15}{0.01}} \\
\multicolumn{1}{|c|}{} 
& DeepSeek-V3 & \dcadp{\textbf{295.28}}{0.19}{4.36} & \dcadp{123.54}{1.27}{0.18} & {\color[HTML]{9B9B9B}\dcadp{8.48}{0.15}{0.01}} \\
\multicolumn{1}{|c|}{} 
& \xname-DS   & \dcadp{\textbf{295.28}}{0.19}{4.36} & \dcadp{\textbf{116.81}}{1.27}{0.17} & \dcadp{\textbf{8.48}}{0.14}{0.01} \\ \hline

\multicolumn{1}{|c|}{\multirow{4}{*}{65nm}} 
& Original    & \dcadp{4021.20}{0.60}{4.61} & \dcadp{1588.68}{4.03}{0.97} & \dcadp{80.64}{0.69}{0.04} \\
\multicolumn{1}{|c|}{} 
& Manual      & {\color[HTML]{9B9B9B}\dcadp{4021.20}{0.60}{4.61}} & {\color[HTML]{9B9B9B}\dcadp{1588.68}{4.03}{0.97}} & {\color[HTML]{9B9B9B}\dcadp{80.64}{0.69}{0.04}} \\
\multicolumn{1}{|c|}{} 
& DeepSeek-V3 & \dcadp{\textbf{2904.12}}{0.59}{3.54} & \dcadp{1195.92}{4.03}{0.78} & {\color[HTML]{9B9B9B}\dcadp{80.64}{0.69}{0.04}} \\
\multicolumn{1}{|c|}{} 
& \xname-DS   & \dcadp{\textbf{2904.12}}{0.59}{3.54} & \dcadp{\textbf{1130.04}}{4.03}{0.75} & \dcadp{\textbf{79.92}}{0.71}{0.04} \\ \hline
\multicolumn{2}{|c|}{Design} & vending\_machine & fft & \textbf{Avg. Impr.} \\ \hline
\multicolumn{2}{|c|}{Metric} & \dcadp{\textbf{Area}}{Delay}{Power} & \dcadp{\textbf{Area}}{Delay}{Power} & \dcadp{\textbf{Area}}{Delay}{Power} \\ \hline

\multicolumn{1}{|c|}{\multirow{2}{*}{SSC}} 
& Original    & \dcadp{240079.30}{7.90}{11.46} & \dcadp{1857805.00}{7.90}{51.12} & - \\
\multicolumn{1}{|c|}{} 
& SymRTLO     & \dcadp{151593.90}{7.90}{8.18} & \dcadp{1471378.00}{8.98}{26.32} & \dcadp{17.58\%}{-4.39\%}{20.12\%} \\ \hline

\multicolumn{1}{|c|}{\multirow{4}{*}{12nm}} 
& Original    & \dcadp{3747.36}{0.23}{17.59} & \dcadp{31271.32}{0.75}{46.68} & - \\
\multicolumn{1}{|c|}{} 
& Manual      & {\color[HTML]{9B9B9B}\dcadp{3747.36}{0.23}{17.59}} & \dcadp{29346.88}{0.81}{37.23} & \dcadp{1.23\%}{-0.27\%}{5.27\%} \\
\multicolumn{1}{|c|}{} 
& DeepSeek-V3 & \dcadp{3608.39}{0.23}{17.59} & {\color[HTML]{9B9B9B}\dcadp{31271.32}{0.75}{46.68}} & \dcadp{11.63\%}{-2.42\%}{12.30\%} \\
\multicolumn{1}{|c|}{} 
& \xname-DS   & \dcadp{\textbf{2966.91}}{0.36}{9.70} & \dcadp{\textbf{27769.19}}{0.79}{42.85} & \dcadp{\textbf{18.12\%}}{-14.79\%}{23.57\%} \\ \hline

\multicolumn{1}{|c|}{\multirow{4}{*}{65nm}} 
& Original    & \dcadp{38829.96}{0.72}{26.75} & \dcadp{304913.90}{2.63}{39.05} & - \\
\multicolumn{1}{|c|}{} 
& Manual      & {\color[HTML]{9B9B9B}\dcadp{38829.96}{0.72}{26.75}} & \dcadp{295164.72}{2.51}{35.67} & \dcadp{0.64\%}{0.91\%}{1.73\%} \\
\multicolumn{1}{|c|}{} 
& DeepSeek-V3 & \dcadp{37383.12}{0.75}{24.96} & {\color[HTML]{9B9B9B}\dcadp{304913.88}{2.63}{39.05}} & \dcadp{11.25\%}{-0.50\%}{10.00\%} \\
\multicolumn{1}{|c|}{} 
& \xname-DS   & \dcadp{\textbf{29798.64}}{1.12}{13.51} & \dcadp{\textbf{262633.70}}{2.73}{35.47} & \dcadp{\textbf{18.93\%}}{-12.12\%}{21.85\%} \\ \hline
\end{tabular}%

}

\label{tab:comp_symrtlo_2}
\end{table*}

\textit{(2) \xname outperforms SymRTLO on more challenging optimization tasks regarding complex and large circuits. }
Since SymRTLO has not open-sourced its optimized RTL code or synthesis scripts, we cannot conduct comparisons under identical standards.
In Table~\ref{tab:comp_symrtlo_2}, we present the self-reported area optimization results from SymRTLO for circuits with complex functionality, manual optimizations, vanilla LLM results, as well as \xname results, using DeepSeek-V3~\cite{deepseekai2025deepseekv3technicalreport} as the backbone model with 5-10-5 search settings. To best align with SymRTLO's synthesis settings, we use a commercial EDA tool, synthesizing circuits into 12nm and 65nm processes, respectively. The experimental results demonstrate that \xname achieves higher area improvement in circuit area compared with SymRTLO in both processes (18.12\% / 18.93\% \vs 17.58\%). However, we emphasize again that differences in process nodes and synthesis scripts can lead to significant variations in PPA outcomes. Since the specific process node and scripts used by SymRTLO were inaccessible to us, achieving a fully fair comparison remains a challenge.

\begin{figure}[h]
  \centering
\includegraphics[width=0.7\linewidth]{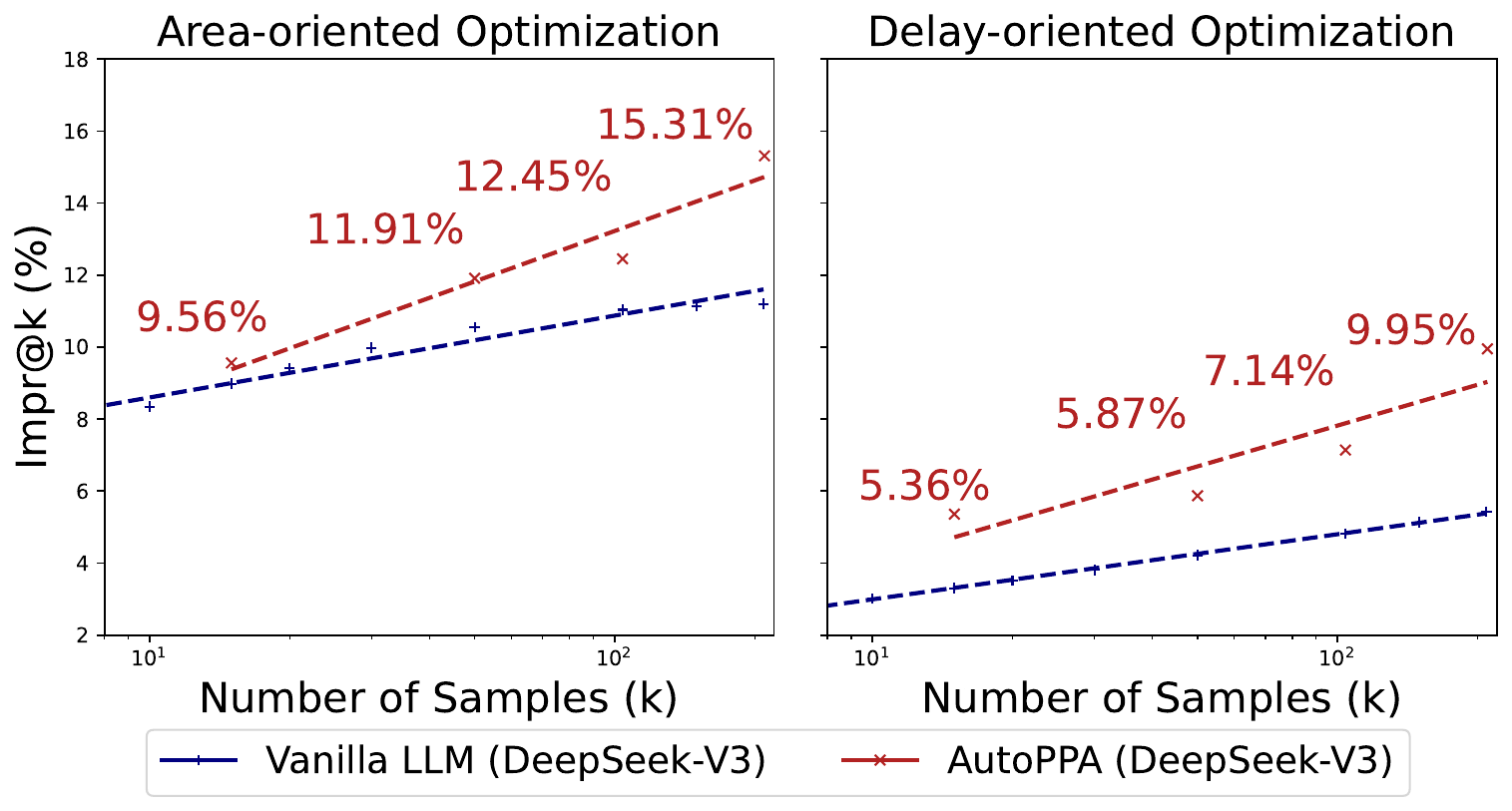}

\caption{Area and Delay improvement comparison with vanilla LLM sampling. \xname yields consistently higher impr@k and better growth
rate than DeepSeek-V3.}
\label{fig:opt_vs_sample}

\end{figure}

\textit{(3) \xname outperforms all baseline LLM methods on both area and delay-oriented optimizations.}
We extend our evaluation to compare our method with baseline LLMs across all \textit{60 circuits} in the complete benchmark, applying settings from 2-3-3 to 5-10-5. Using Qwen2.5-7B series models and DeepSeek-V3 as backbone models for search, we compare against the language models listed in Table~\ref{tab:baseline_llms} targeting either area or delay optimization. The results are reported in Tables~\ref{tab:main_area_delay} and illustrated in Figure~\ref{fig:opt_vs_sample}.

\begin{table*}[t]
\centering
\caption{Comparison of \uline{area-oriented} and \uline{delay-oriented} optimization with vanilla LLMs across \uline{60} circuits.}

\resizebox{\linewidth}{!}{%
\begin{tabular}{|cc|c|c|c|c|}
\hline
\multicolumn{6}{|c|}{\textbf{Area-oriented Optimization}} \\ \hline
\multicolumn{2}{|c|}{} &
  \multicolumn{1}{c|}{$Impr@15$ (\%)} &
  \multicolumn{1}{c|}{$Impr@50$ (\%)} &
  \multicolumn{1}{c|}{$Impr@104$ (\%)} &
  $Impr@210$ (\%) \\ \cline{3-6}
\multicolumn{2}{|c|}{\multirow{-2}{*}{Method}} &
  \multicolumn{1}{c|}{\dcadp{\textbf{Area}}{Delay}{Power}} &
  \multicolumn{1}{c|}{\dcadp{\textbf{Area}}{Delay}{Power}} &
  \multicolumn{1}{c|}{\dcadp{\textbf{Area}}{Delay}{Power}} &
  \dcadp{\textbf{Area}}{Delay}{Power} \\ \hline
\multicolumn{1}{|c|}{} & HaVen &
  \multicolumn{1}{c|}{\adp{1.48}{0.12}{1.33}} &
  \multicolumn{1}{c|}{\adp{3.46}{0.33}{3.02}} &
  \multicolumn{1}{c|}{\adp{4.82}{0.51}{4.14}} &
  \adp{5.57}{0.60}{4.77} \\
\multicolumn{1}{|c|}{\multirow{-2}{*}{\begin{tabular}[c]{@{}c@{}}RTL-Specific\\Models\end{tabular}}} &
  CodeV &
  \multicolumn{1}{c|}{\adp{1.61}{0.47}{1.35}} &
  \multicolumn{1}{c|}{\adp{3.53}{1.06}{3.10}} &
  \multicolumn{1}{c|}{\adp{4.75}{1.63}{4.36}} &
  \adp{5.45}{2.27}{5.32} \\ \hline
\multicolumn{1}{|c|}{Reasoning} & DS-R1-Dist &
  \multicolumn{1}{c|}{\adp{3.86}{0.55}{3.74}} &
  \multicolumn{1}{c|}{\adp{6.27}{1.21}{6.01}} &
  \multicolumn{1}{c|}{\adp{8.11}{1.90}{7.78}} &
  \adp{9.55}{2.76}{9.22} \\ \hline
\multicolumn{1}{|c|}{} & Qwen-Coder &
  \multicolumn{1}{c|}{\adp{5.52}{1.29}{4.79}} &
  \multicolumn{1}{c|}{\adp{8.00}{1.97}{6.90}} &
  \multicolumn{1}{c|}{\adp{10.10}{2.87}{8.74}} &
  \adp{12.37}{3.97}{10.60} \\
\multicolumn{1}{|c|}{\multirow{-2}{*}{\begin{tabular}[c]{@{}c@{}}General\\Models\end{tabular}}} &
  DeepSeek-V3 &
  \multicolumn{1}{c|}{\adp{9.95}{1.63}{8.98}} &
  \multicolumn{1}{c|}{\adp{11.40}{1.85}{10.55}} &
  \multicolumn{1}{c|}{\adp{11.97}{2.09}{11.04}} &
  \adp{12.29}{2.36}{11.20} \\ \hline
\multicolumn{1}{|c|}{} & \xname-Qw &
  \multicolumn{1}{c|}{\adp{5.41}{2.12}{4.98}} &
  \multicolumn{1}{c|}{\adp{9.97}{3.25}{8.83}} &
  \multicolumn{1}{c|}{\adp{11.13}{2.94}{9.50}} &
  \adp{14.31}{4.87}{11.05} \\
\multicolumn{1}{|c|}{\multirow{-2}{*}{Ours}} &
  \xname-DS &
  \multicolumn{1}{c|}{\adp{10.84}{1.66}{\textbf{9.56}}} &
  \multicolumn{1}{c|}{\adp{13.44}{2.57}{\textbf{11.91}}} &
  \multicolumn{1}{c|}{\adp{13.58}{2.05}{\textbf{12.45}}} &
  \adp{15.97}{0.43}{\textbf{15.31}} \\ \hline

\multicolumn{6}{|c|}{\textbf{Delay-oriented Optimization}} \\ \hline
\multicolumn{2}{|c|}{} &
  \multicolumn{1}{c|}{$Impr@15$ (\%)} &
  \multicolumn{1}{c|}{$Impr@50$ (\%)} &
  \multicolumn{1}{c|}{$Impr@104$ (\%)} &
  $Impr@210$ (\%) \\ \cline{3-6}
\multicolumn{2}{|c|}{\multirow{-2}{*}{Method}} &
  \multicolumn{1}{c|}{\dcadp{\textbf{Delay}}{Area}{Power}} &
  \multicolumn{1}{c|}{\dcadp{\textbf{Delay}}{Area}{Power}} &
  \multicolumn{1}{c|}{\dcadp{\textbf{Delay}}{Area}{Power}} &
  \dcadp{\textbf{Delay}}{Area}{Power} \\ \hline
\multicolumn{1}{|c|}{} & HaVen &
  \multicolumn{1}{c|}{\dap{0.46}{0.93}{1.00}} &
  \multicolumn{1}{c|}{\dap{1.38}{1.89}{1.88}} &
  \multicolumn{1}{c|}{\dap{2.11}{2.60}{2.33}} &
  \dap{2.97}{3.30}{2.49} \\ 
\multicolumn{1}{|c|}{\multirow{-2}{*}{\begin{tabular}[c]{@{}c@{}}RTL-Specific\\Models\end{tabular}}} &
  CodeV &
  \multicolumn{1}{c|}{\dap{1.98}{2.50}{1.92}} &
  \multicolumn{1}{c|}{\dap{2.49}{3.36}{2.49}} &
  \multicolumn{1}{c|}{\dap{2.39}{3.34}{2.80}} &
  \dap{1.69}{2.61}{3.15} \\ \hline
\multicolumn{1}{|c|}{Reasoning} & DS-R1-Dist &
  \multicolumn{1}{c|}{\dap{1.87}{2.32}{1.51}} &
  \multicolumn{1}{c|}{\dap{2.47}{3.00}{2.52}} &
  \multicolumn{1}{c|}{\dap{2.48}{3.06}{3.02}} &
  \dap{2.00}{2.56}{3.33} \\ \hline
\multicolumn{1}{|c|}{} & Qwen-Coder &
  \multicolumn{1}{c|}{\dap{3.61}{4.51}{2.76}} &
  \multicolumn{1}{c|}{\dap{4.02}{5.11}{4.61}} &
  \multicolumn{1}{c|}{\dap{4.56}{5.89}{6.11}} &
  \dap{5.83}{7.49}{7.74} \\ 
\multicolumn{1}{|c|}{\multirow{-2}{*}{\begin{tabular}[c]{@{}c@{}}General\\Models\end{tabular}}} &
  DeepSeek-V3 &
  \multicolumn{1}{c|}{\dap{1.65}{2.90}{4.20}} &
  \multicolumn{1}{c|}{\dap{2.11}{3.44}{5.15}} &
  \multicolumn{1}{c|}{\dap{2.48}{3.96}{5.67}} &
  \dap{2.90}{4.49}{6.37} \\ \hline
\multicolumn{1}{|c|}{} & AutoPPA-Qw &
  \multicolumn{1}{c|}{\dap{3.10}{4.62}{4.65}} &
  \multicolumn{1}{c|}{\dap{4.82}{5.52}{\textbf{6.15}}} &
  \multicolumn{1}{c|}{\dap{9.75}{11.79}{\textbf{9.80}}} &
  \dap{8.00}{10.87}{\textbf{11.28}} \\ 
\multicolumn{1}{|c|}{\multirow{-2}{*}{Ours}} &
  AutoPPA-DS &
  \multicolumn{1}{c|}{\dap{1.09}{1.54}{\textbf{5.36}}} &
  \multicolumn{1}{c|}{\dap{0.73}{1.74}{5.87}} &
  \multicolumn{1}{c|}{\dap{2.10}{2.44}{7.14}} &
  \dap{6.37}{7.78}{9.95} \\ \hline

\end{tabular}
}
\label{tab:main_area_delay}

\end{table*}

The results in Tables~\ref{tab:main_area_delay} demonstrate that our method achieves the best optimization among models of similar size for both area and delay optimization objectives. Importantly, our method shows consistent improvement as the search budget increases. Figure~\ref{fig:opt_vs_sample} further illustrates the comparison between \xname and the vanilla DeepSeek-V3 for optimization:
as the number of samples grows, \xname yields consistently higher impr@k and better growth rate than DeepSeek-V3.
At the largest search budget, our method outperforms vanilla LLM by 4.11\% for area-oriented optimization and 3.58\% for delay-oriented optimization.

Notably, when targeting delay optimization, utilizing the smaller Qwen2.5-7B series models as a search backbone yields better results than using DeepSeek-V3. We attribute this to Qwen2.5-7B's superior baseline performance on delay optimization tasks compared to DeepSeek-V3. Despite being distilled from DeepSeek-R1 and fine-tuned on Qwen2.5-Math-7B, DS-R1-Distill's optimization capability falls short of the Qwen2.5-Coder-Instruct model with the same architecture and parameter scale, suggesting that distillation of reasoning models leads to performance degradation on specific tasks. HaVen and CodeV, despite being specialized models for RTL generation, demonstrate the lowest optimization on the RTL optimization task. This highlights the importance of multi-task fine-tuning for promoting domain-specific capabilities.

\textbf{\xname is effective on various synthesis settings.} 
Combining results from Table~\ref{tab:comp_with_rtlrewriter} and Table~\ref{tab:comp_symrtlo_2}, \xname shows significant PPA improvements on both open-source and commercial EDA toolchains. This result highlights the wide applicability of \xname.
Notably, \xname achieves significant area optimization across 12nm to 65nm process nodes in 2 large, practical circuits from RTLRewriter Benchmark (\textit{fft} and \textit{vending\_machine}), reaching 11.20\% and 20.83\% in the 12nm process, and 13.87\% and 23.26\% in the 65nm process.
We observe that, although the same circuit exhibits similar optimization trends across different process nodes, there are noticeable differences in the specific optimization ratios. This underscores the importance of employing unified synthesis toolchains and process nodes to ensure fair evaluation across different works.
To the best of our knowledge, we are the first work to evaluate LLM-based RTL code PPA optimization across multiple EDA tools (\code{black}{SiliconCompiler} and a commercial tool) and processes (from 12nm, 45nm, to 65nm).

\textbf{\xname's rule library surpasses manually crafted rule libraries.} 
We conduct an experiment to compare the effectiveness of optimization rules automatically derived from our \textit{explore-evaluate-induce} (\EEI) process against those manually summarized by Verilog engineers. The comparison uses an identical 3-5-4 search setting, employing Qwen2.5 series models as the search backbone for area-oriented optimization on all 60 circuits. The only difference is the rule library used during optimization. Table~\ref{tab:compair_with_manual_lib} presents the experimental results. The scalability of our \EEI process generates significantly more rules (101,987 rules) compared to 16 hours of manual efforts (12 rules), resulting in superior optimization outcomes.

\begin{table}[t]
\centering
\caption{\uline{Area-oriented} optimization comparison between \xname's learned rule library with the manually constructed one, tested with \xname-Qw.}

\resizebox{0.7\linewidth}{!}{
\begin{tabular}{|l|c|}
\hline
\multirow{2}{*}{Settings}  & $Impr@50$ (\%) \\ \cline{2-2} 
\multicolumn{1}{|c|}{} & \dcadp{\textbf{Area}}{Delay}{Power} \\ \hline
\xname-Qw with \EEI rules (ours)   & {\dcadp{\textbf{8.83\%}}{3.25\%}{9.97\%}} \\
with manual rules & {\dcadp{6.56\%}{0.80\%}{5.46\%}}  \\ \hline
\end{tabular}}
\label{tab:compair_with_manual_lib}

\end{table}

\textbf{Ablation Study.} To validate the effectiveness of each component in our proposed approach, we conduct a series of ablation studies. Our complete design incorporates LLM-generated rule \textbf{speculation}, rule library \textbf{retrieval} with these speculative results, \textbf{adaptation} of retrieved rules, and the \textbf{search} framework. We perform experiments for ablations under equal RTL code sampling overhead. Table~\ref{tab:ablation} presents the experimental results, which demonstrate the efficacy of each component in our design. The performance degradation observed in each ablation setting validates our integrated approach for optimal performance.

\begin{table}[h]
\centering
\caption{Ablation on \xname's adaptive rule-based PPA optimization framework on \uline{area-oriented} optimizations.}

\resizebox{0.6\linewidth}{!}{%
\begin{tabular}{|l|c|}
\hline
\multirow{2}{*}{Settings}  & $Impr@50$ (\%) \\ \cline{2-2}
\multicolumn{1}{|c|}{}       & \dcadp{\textbf{Area}}{Delay}{Power} \\ \hline
\xname-Qw (ours)                   & {\adp{9.97}{3.25}{\textbf{8.83}}}     \\
w/o Search  & {\adp{7.94}{3.14}{7.83}}               \\
w/o Adapt& {\adp{7.62}{2.58}{6.96}}               \\
w/o Retrieve, Adapt & {\adp{5.67}{2.41}{5.34}}               \\
w/o Speculate, Retrieve, Adapt                      & {\adp{5.84}{1.95}{6.39}}               \\
 \hline
\end{tabular}%
}

\label{tab:ablation}
\end{table}

\section{Related work}

\textbf{LLM-based RTL Generation.} Recent works leverage Large Language Models (LLMs) to assist circuit design flows. Frameworks such as Chip-Chat~\cite{chipchat}, OriGen~\cite{cui2024origen}, RTLFixer~\cite{tsai2024rtlfixer}, and VerilogCoder~\cite{verilogcoder} use self-reflection~\cite{reflexion} or multi-agent systems for RTL generation. Others, including VeriGen~\cite{thakur2024verigen}, RTLCoder~\cite{liu2024rtlcoder}, BetterV~\cite{pei2024betterv}, CodeV~\cite{zhao2025codev,zhu2025codevr1}, and HaVen~\cite{yang2025haven}, build domain-specific datasets and fine-tune LLMs to improve generation quality. However, these efforts primarily address circuit generation~\cite{liu2025openllm,Pinckney2025revisitingverilogeval} rather than circuit optimization challenges.

\textbf{LLM-based RTL Optimization.} Circuit optimization differs from generation tasks as it requires maintaining functional equivalence while understanding the complex relation among hardware code, circuit structure, and post-synthesis QoR metrics.
Some approaches attempt to address circuit optimization directly. ChipGPT \cite{chipgpt} and VeriPPA~\cite{thorat2024advancedlargelanguagemodel} feed post-synthesis PPA metrics back to LLMs and request optimization directly.
RTLRewriter~\cite{yao2024rtlrewriter} builds a manual knowledge base, retrieves information from code and diagrams, and applies Monte Carlo Tree Search (MCTS) for rewriting. SymRTLO~\cite{wang2025symrtlo} formats manually collected optimization materials~\cite{Knoop1994deadcode,Pasko1999subexpelim} for retrieval during iterative refinement. These approaches depend on a manually constructed knowledge base with limited entries and task-specific optimization capabilities.

\section{Conclusion}

In this paper, we present~\xname, a fully automated PPA optimization framework comprising an automatically generated optimization rule library and an enhanced optimization search method.
To mitigate the scarcity of optimization rules and the challenges in rule induction, we develop a rule library learning framework based on an \textit{Explore-Evaluate-Induce} workflow. 
To address the inefficiency in applying abstract optimization rules to concrete designs, we propose an Adaptive Retrieval Augmented Optimization (ARAO) mechanism integrated with a multi-step search strategy.
Experimental results demonstrate that~\xname surpasses baseline LLM methods, achieves a 19.25\% improvement in area optimization over manual methods and a 7.56\% improvement over RTLRewriter.

\bibliographystyle{splncs04}
\bibliography{references}

\end{document}